\documentclass[letterpaper, 10 pt, conference]{ieeeconf}  % Comment this line out
\usepackage[pdftex]{graphicx}
\graphicspath{{resources/}}

\usepackage{ifthen}

\usepackage{amssymb}

\usepackage[T1]{fontenc}
\usepackage[utf8]{inputenc}
\usepackage{xspace}  
\usepackage{cite}  
\usepackage{ifthen}

\usepackage{xcolor}
\usepackage[normalem]{ulem}

\newboolean{showcomments}
\setboolean{showcomments}{true}
\ifthenelse{\boolean{showcomments}}
	{\newcommand{\nb}[3]{
		{\colorbox{#2}{\bfseries\sffamily\scriptsize\textcolor{white}{#1}}}
		{\textcolor{#2}{\sf\small$\blacktriangleright$\textit{#3}$\blacktriangleleft$}}}
	 }
	{\newcommand{\nb}[3]{}
	 }

\makeatletter
\newcommand{\camelhyph}[1]{\@fterfirst\c@amelhyph#1\relax }
\def\@fterfirst #1#2{#2#1}
\def\c@amelhyph #1{%
 \ifthenelse{\equal{#1}\relax}{}{%  Do nothing if the end has been reached
   \ifnum`#1<91 \-#1\else#1\fi%     Check whether #1 is an uppercase letter,
  \expandafter\c@amelhyph%    %     insert \c@amelhyph again.
}}
 
\makeatother

\newcommand{\diaspec}{Dia\-Spec\xspace}
\newcommand{\ct}[1]{\texttt{\camelhyph{#1}}}
\newcommand{\ie}{\emph{i.e.,}}
\newcommand{\eg}{\emph{e.g.,}}
\newcommand{\etc}{\emph{etc}}
\newcommand{\etal}{\emph{et al.}}

\title{Using the \diaspec{} design language and compiler to develop
  robotics systems}

\author{%
  \parbox{2.4 in}{\centering Damien Cassou\\
    Software Architecture Group, HPI\\
    University of Potsdam, Germany\\%
    {\tt\small damien.cassou@hpi.uni-potsdam.de}}
  \hspace*{ 0.1 in}
  \parbox{2.35 in}{ \centering Serge Stinckwich\\
UMI 209 UMMISCO\\IRD/IFI/Vietnam National University
    {\tt\small serge.stinckwich@ird.fr}}
  \hspace*{ 0.1 in}
  \parbox{1.9 in}{ \centering Pierrick Koch\\
UMR 6072 GREYC\\Université de Caen-Basse Normandie/CNRS/ENSICAEN\\
    {\tt\small pierrick.koch@unicaen.fr}}
}

\usepackage{listings}

\lstdefinelanguage{diaspec}
{morekeywords={import, entity, attribute, extends, source, action, controller,
    context, from, on, void, Integer, Boolean, Float, String, enumeration,
    structure, indexed, by, as, include, when, get, no, maybe, always,
    publish, interaction, required, provided, abstract, do, any},%
  sensitive=true,
  morecomment=[l]{//},
  morecomment=[s]{/*}{*/},
}

\usepackage[pdftex, colorlinks=true, urlcolor=black, linkcolor=black]{hyperref}
\begin{document}

\maketitle
\thispagestyle{empty}
\pagestyle{empty}

%%%%%%%%%%%%%%%%%%%%%%%%%%%%%%%%%%%%%%%%%%%%%%%%%%%%%%%%%%%%%%%%%%%%%%%%%%%%%%%%
\begin{abstract}

  A Sense/Compute/Control (SCC) application is one that interacts with
  the physical environment. Such applications are pervasive in domains
  such as building automation, assisted living, and autonomic
  computing. Developing an SCC application is complex because: (1) the
  implementation must address both the interaction with the
  environment and the application logic; (2) any evolution in the
  environment must be reflected in the implementation of the
  application; (3) correctness is essential, as effects on the
  physical environment can have irreversible consequences.

  The SCC architectural pattern and the \diaspec{} domain-specific
  design language propose a framework to guide the design of such
  applications. From a design description in \diaspec{}, the
  \diaspec{} compiler is capable of generating a programming framework
  that guides the developer in implementing the design and that
  provides runtime support. In this paper, we report on an experiment
  using \diaspec{} (both the design language and compiler) to develop
  a standard robotics application. We discuss the benefits and
  problems of using \diaspec{} in a robotics setting and present some
  changes that would make \diaspec{} a better framework in this
  setting.

\end{abstract}

%!TEX root=dslrob.tex
\section{Introduction}

A Sense/Compute/Control (SCC) application is one that interacts with
the environment~\cite{Tayl09a}. The SCC architectural pattern
guides the description of SCC applications and involves four kinds of
components, organized into layers~\cite{Cass11a,Edwar09a}: (1)
\emph{sensors} at the bottom, which obtain information about the
environment; (2) then \emph{context operators}, which process this
information; (3) then \emph{control operators}, which use this refined
information to control (4) \emph{actuators} at the top, which finally
impact the environment. A robotics application is a kind of SCC
application where the environment is composed of a robot
(sensors/actuators/body, control architecture, \etc{}) and the robot's
neighborhood (the walls, ground, people, \etc{})~\cite{Sicil08a}. As
noticed by Taylor \etal{}~\cite{Tayl09a}, the Sense/Plan/Act
architecture~\cite{Sicil08a}, widely used in robotics, closely
resembles the SCC architectural pattern.

\diaspec{} is a domain-specific design language dedicated to
describing SCC applications~\cite{Cass09b,Cass11a}. From such a design
description, the \diaspec{} compiler produces a dedicated Java
programming framework that is both \emph{prescriptive} and
\emph{restrictive}: it is prescriptive in the sense that it guides the
developer, and it is restrictive in the sense that it limits the
developer to what the design description allows. By separating
application logic (implemented by the developers) and runtime support
(generated in the programming framework), \diaspec{} facilitates the
design, implementation and evolution of SCC applications.

%\damien{Talk about problems in the robotics domain: ad-hoc solutions, hard to reuse, hard to adapt to new environments...}

\subsection*{Contributions}

Our contributions are as follows:

\begin{itemize}
\item \emph{A report} on an experiment of designing and implementing a
  standard robotics application in the SCC architectural pattern with  the \diaspec{} domain-specific design language and framework
  (Sections~\ref{sec:designing} and~\ref{sec:implementing}). This
  report includes detailed instructions and guidelines to allow
  further experiments.
\item \emph{A discussion} of the benefits and problems of using
  \diaspec{} in a robotics setting (Section~\ref{sec:discussing}).
  This discussion includes a list of changes to \diaspec{} that would
  make it a better framework for developing new robotics applications.
\end{itemize}

We finally highlight some related works and conclude in
sections~\ref{sec:related} and~\ref{sec:conclusion}.

%%% LocalWords:  SCC

%!TEX root=dslrob.tex
\section{Designing a robotics application}
\label{sec:designing}

In this section we first explain how to decompose a robotics
application in \diaspec{} component types. Then we present a case
study that we use as an example of how to describe a robotics
application with \diaspec{}.

In the rest of this paper we use
ROS\footnote{\url{http://www.ros.org/wiki/}} as the underlying
middleware for our case study. We believe it is a good choice as ROS
is becoming a standard within the robotics community. It is important
to note however that our approach and \diaspec{} are independent of
any middleware.

\subsection{Decomposing}

Designing an application with \diaspec{} requires a decomposition in
layers. Each layer corresponds to a separate type of component:

\begin{itemize}
\item A \emph{sensor} sends information sensed from the environment to
  the context operator layer through data \emph{sources}. A sensor can
  both push data to context operators and respond to context operator
  requests. We use the term ``sensor'' both for entities that actively
  retrieve information from the environment, such as system probes,
  and entities that store information previously collected from the
  environment, such as structured information coming from the
  middleware.
\item A \emph{context operator} refines (aggregates and interprets)
  the information given by the sensors. Context operators can push
  data to other context operators and to control operators. Context
  operators can also respond to requests from parent context
  operators.
\item A \emph{control operator} transforms the information given by
  the context operators into orders for the actuators.
\item An \emph{actuator} triggers actions on the environment.
\end{itemize}

The following details the steps to follow to decompose a robotics
application into these component types.

\paragraph*{Reusing existing components}
In the presence of a previous application developed with \diaspec{},
it is possible and advisable to reuse as much components as possible.
Depending on the amount of reused components this can have a huge
impact on the application of the other steps.
% shouldn't talk about middleware reuse here as it is discussed below:
% It is similarly important to reuse as much building blocks as
% possible from the underlying middleware. For example ROS provides
% high-level components and algorithms that are better reused than
% reimplemented.

\paragraph*{Listing capabilities}
Each robot comes with its own set of capabilities (\eg{} sensing
motion and projecting light). A developer should map these
capabilities to sensor sources and actuator actions. A developer
should then group related sources and actions inside \emph{entity
  classes} (\eg{} a camera providing a picture source and zooming
action). Beside sources and actions, an entity class may also have
\emph{attributes} to characterize its instances (\eg{} resolution,
accuracy and status). In the presence of a high-level middleware (such
as ROS), it can be useful to also map capabilities of the middleware
into sources and actions (\eg{} a mapping or a localization
capability).

\paragraph*{Identifying main context operators}
The next step of the decomposition in components is the identification
of the main high-level pieces of information required by the
application. A developer should map these pieces of information to
context operators and use them as input to control operators.

\paragraph*{Decomposing into lower-level pieces}
Then, a developer must identify lower-level context operators that act
as input sources for the higher-level ones. This decomposition is
typically done in several steps, each step slightly lowering the level
of previously identified context operators. This decomposition ends
when each identified context operator can directly take its input from
a set of sensor sources.

\paragraph*{Identifying control operators}
From the high-level context operators a developer has to derive a set
of control operators that will send orders to actuators. Because a
developer can not reuse the code of a control operator in another part
of the application, it is important that this code is as simple as
possible. If there is opportunity for reuse, the code should be moved
to a new context operator.

\paragraph*{Identifying data types}
While proceeding with the above steps it is also necessary to define
data types. A developer then use these types to describe entity
sources, context operators, and parameters of actuator actions. A data
type is either primitive (\eg{} integer, boolean and float), an
enumeration of names (\eg{} a luminosity can either be low, normal or
high), or a structure (\eg{} a coordinate with x and y fields). An
important question arises in the presence of a high-level middleware
(such as ROS): should the types of the application be the types
provided by the middleware or should the application define new types.
The former solution is easier to use whereas the latter provides more
decoupling. A general principle is to provide new types when their
transcription in \diaspec{} is straightforward (\eg{} a coordinate)
and to reuse the middleware types otherwise (\eg{} ROS defines a
``twist'' data type that is complex enough to not be reimplemented).

\subsection{Case Study}

As a running example, we present an application that is typical of the
robotics domain. In this application, a robot evolves in an unknown
environment and has two modes: \emph{random} and \emph{exploration}.
In the random mode, the robot goes straight and when an obstacle is
close enough turns before going straight again. In the exploration
mode, the robot goes to unvisited locations with the goal to visit as
much as possible from the neighborhood. The current mode can be
changed at anytime by an operator through a graphical interface. In
both modes, the robot turns on an embedded projector and takes
pictures when it is in a zone with obstacles.

\begin{figure}
  \centering
  \includegraphics[width=.7\linewidth]{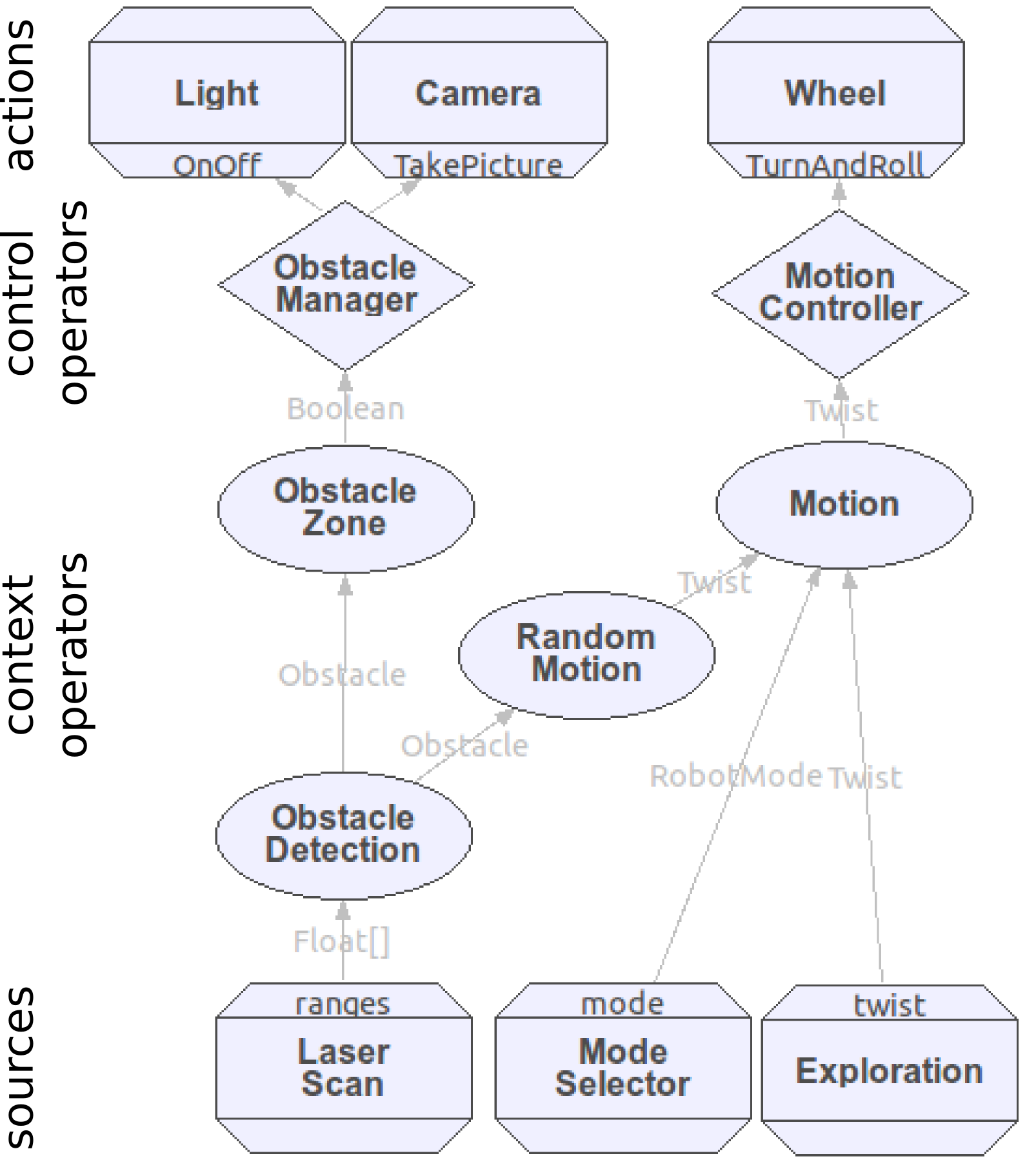}
  \caption{The case study decomposed into the different type of
    components of \diaspec{}}
\label{fig:diaspec-graph}
\end{figure}

Let us now discuss the above steps in the context of this case
study (Figure~\ref{fig:diaspec-graph} represents the result).

\paragraph*{Reusing existing components}
We assume no previous \diaspec{} application and thus no \diaspec{}
component to reuse.

\paragraph*{Listing capabilities}
The Bosch robotics research group develops an \emph{exploration}
capability\footnote{\url{http://www.ros.org/wiki/explore}} based on a
well-known frontier-based exploration algorithm~\cite{Yamau98a}. In
this algorithm the exploration is composed of two steps: a motion
toward a location and a new observation of the environment at this
location. The location is chosen among a set of candidate locations on
the frontier between explored and unexplored space. This capability is
exactly what we need for the exploration mode of the robot. Our robot
comes with a range-finder type laser scanner, a light projector, a
camera, and a set of wheels. From all these capabilities, we identify:

\begin{itemize}
\item a \ct{LaserScan} entity class with a \ct{ranges} source
  providing laser ranges from the sensor;
\item A \ct{ModeSelector} entity that provides a graphical interface
  for the operator to choose the current mode of the robot;
\item an \ct{Exploration} entity that provides a source of twists for
  the robot;
\item a \ct{Light} and \ct{Camera} entities that respectively
  enlighten the neighborhood and take pictures on request;
\item a \ct{Wheel} entity that can turn or roll on request;
\end{itemize}

\paragraph*{Identifying main context operators}
The most important activity of our robot is to move. Therefore we
introduce a \ct{Motion} context operator that produces a twist,
representing the motion of the robot. Because our robot takes pictures and
turns on its projector when it is in a zone with obstacles, we
introduce an \ct{ObstacleZone} context operator that indicates whether
or not some obstacles are in the neighborhood.

\paragraph*{Decomposing into lower-level pieces}
The \ct{Motion} context operator produces a twist based on which mode
is selected and on the twist values coming from both modes. The
selected mode is directly provided by the \ct{ModeSelector} entity. We
introduce a \ct{RandomMotion} context operator that produces twists
for the random mode. The twist for the exploration mode is directly
provided by the \ct{Exploration} entity. Both the \ct{ObstacleZone}
and \ct{RandomMotion} context operators need the information about
nearby obstacles. We thus introduce the \ct{ObstacleDetection} context
operator to indicate the proximity of an obstacle.

\paragraph*{Identifying control operators}
The \ct{MotionController} control operator takes information from the
\ct{Motion} context operator and transmits this information to the
\ct{Wheel} entity. The \ct{ObstacleManager} control operator takes
information from the \ct{ObstacleZone} context operator and triggers the
light and takes a picture with the camera.

\paragraph*{Identifying data types}
We have already seen that our application uses the notion of twist to
indicate motion. A developer can define a twist as a pair of vectors
which represent the linear and angular velocity. The robot current
mode is represented as an enumeration of the \ct{RANDOM} and
\ct{EXPLORATION} names. The \ct{ObstacleDetection} context operator
provides an \ct{Obstacle} data type containing both a boolean to
indicate if an obstacle is in front of the robot and a set of float
numbers (the ranges) as provided by the laser scan giving details
about the neighborhood.

\subsection{Describing with \diaspec{}}

Once a developer decomposed the application using the different
component types, the transcription to the \diaspec{} design language
is straightforward. Listing~\ref{listing:design} gives an extract of
the case study transcription.

\lstinputlisting[float,language=diaspec, numbers=left,
breakatwhitespace=true,%
caption={An extract of the description of the robotics application
  with the \diaspec{} design language}%
,label={listing:design}]%
{code/design.diaspec}

In this listing, the \ct{entity}, \ct{context}, and \ct{controller}
keywords are respectively used to introduce a new entity class, a new
context operator, and a new control operator. For this application, we
decide to reuse the \ct{Twist} data type of the ROS middleware which
Listing~\ref{listing:design} illustrates in line~\ref{design:import}.

% interaction contracts: Additionally to the description of an
% operator's input sources, the \diaspec{} language allows each
% operator to describe a set of \emph{interaction contracts} that
% coordinate these input sources. An interaction contract is a tuple
% that describes an \emph{activation condition} indicating which input
% sources can activate the operator and a \emph{reaction} indicating
% what to do as a result of the activation
% condition.\footnote{Previous work~\cite{Cass11a} also includes in
% the tuple a set of \emph{data requirements} indicating the
% additional input sources an operator can use for a particular
% activation condition. This is not used in this case study.} The
% \ct{interaction} keyword is used to introduce an interaction
% contract. For example, the \ct{ObstacleDetection}'s interaction
% contract (Listing~\ref{listing:design},
% line~\ref{design:interaction}) indicates that this context operator
% always publishes a data when it receives information from the laser
% scan.

In this section we saw how to design a robotics application using
the SCC architectural pattern and the \diaspec{} design language. Both
the pattern and the language help decomposing an application in well
defined components. Both make it easy to reuse as much as possible
from the underlying middleware and existing applications. In the next
section we discuss how to implement a robotics application with our
approach.

%%% Local Variables:
%%% mode: latex
%%% coding: utf-8
%%% TeX-master: "dslrob"
%%% TeX-PDF-mode: t
%%% ispell-local-dictionary: "english"
%%% End:

% LocalWords:  boolean

%!TEX root=dslrob.tex

\section{Implementing a robotics application}
\label{sec:implementing}

\begin{figure}
  \centering
  \includegraphics[width=.8\linewidth]{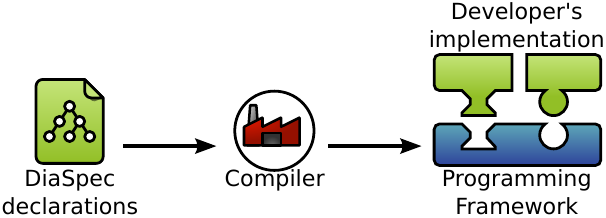}
  \caption{Overview of the \diaspec{} development process}
  \label{fig:diaspec-process}
\end{figure}

The \diaspec{} compiler generates a programming framework with respect
to a set of declarations for entity classes, context operators and
control operators (Figure~\ref{fig:diaspec-process}). For each
component declaration (entity or operator) the compiler generates an
abstract class. The abstract methods in this class represent code to
be provided by the developer (\emph{hole} in
Figure~\ref{fig:diaspec-process}), to allow him to program the
application logic (\eg{} to trigger an entity action) (\textit{bump}
in Figure~\ref{fig:diaspec-process}).

Implementing a \diaspec{} component is done by \textit{sub-classing}
the corresponding generated abstract class. In doing so, the developer
is required to implement each abstract method. The developer writes
the application code in subclasses, not in the generated abstract
classes. This strategy contrasts with generating incomplete source
code to be filled by the developer. As a result, in our approach, one
can change the \diaspec{} declarations and generate a new programming
framework without overriding the developer's code. The mismatches
between the existing code and the new programming framework are
revealed by the Java compiler. To facilitate the implementation
process, most Java IDEs are capable of generating class templates
based on super abstract classes.

In this section, we give an overview of how to implement some parts of
the case study. For a more detailed description, we refer to our
previous works~\cite{Cass09b,Cass11a,Cass11b}.

\subsection{Implementing an operator}

For each context or control operator a dedicated abstract class is
generated in the programming framework. For each input source of this
operator the generated abstract class contains an \emph{abstract
  method} and a corresponding \emph{calling method}. The abstract
method is to be implemented by the developer while the calling method
is used by the framework to call the implementation of the abstract
method with the expected arguments.

% The signature of each abstract method is directly derived from the
% interaction contract: the name of the method is derived from the
% activation condition, the return type is derived from the reaction,
% and the parameters are derived from the activation condition and the
% reaction.

Listing~\ref{listing:contextop-implem} presents a possible Java
implementation of the \ct{RandomMotion} context operator. The
\ct{onObstacleDetection} method is declared abstract in the
\ct{AbstractRandomMotion} generated super class.

\lstinputlisting%
[float,language=java,%
caption={A developer-supplied Java implementation of the
  \texttt{Random\-Motion} context operator described in
  Listing~\ref{listing:design}. The \texttt{Abstract\-Random\-Motion}
  super class is automatically generated into the programming
  framework},%
label={listing:contextop-implem}]%
{code/RandomMotion.java}

Because an operator only manipulates input sources to produce a
result, its implementation is independent of any robotics software
framework. This facilitates operator reuse for different applications
and robots.

\subsection{Implementing an entity}

Contrary to operators which are dedicated to the application logic, an
entity is at the border between the application and its environment
(\eg{} the middleware and robot hardware). Implementing an entity thus
requires some knowledge of the underlying middleware or hardware.

Listing~\ref{listing:laserscan-implem} presents a possible Java
implementation of the \ct{LaserScan} entity class for the ROS
middleware. When the middleware publishes a new laser scan message,
this message is automatically received by the \ct{RosLaserScan}
instance through the ROS \ct{MessageListener} interface.

\lstinputlisting%
[float,language=java,%
caption={A developer-supplied Java implementation of the
  \texttt{LaserScan} entity class described in
  Listing~\ref{listing:design}, line~\ref{design:laserscan-b}. The
  \texttt{Abstract\-Laser\-Scan} super class is automatically
  generated into the programming framework},%
label={listing:laserscan-implem}]%
{code/RosLaserScan.java}

Listing~\ref{listing:light-implem} presents a possible Java
implementation of the \ct{Light} entity class for the ROS middleware.
The constructor receives a ROS \ct{publisher} as a parameter which
allows the entity implementation to send commands to the robot through
the middleware.

\lstinputlisting%
[float,language=java,%
caption={A developer-supplied Java implementation of the
  \texttt{Light} entity class described in
  Listing~\ref{listing:design}, line~\ref{design:light-b}. The
  \texttt{Abstract\-Light} super class is automatically generated into
  the programming framework},%
label={listing:light-implem}]%
{code/RosLight.java}

\subsection{Deploying an application}

Deploying an application requires writing a deployment script in Java.
To do this, a developer creates a new Java class by sub-classing the
abstract class \ct{MainDeploy} generated in the programming framework.
By doing so the developer is required to implement one abstract method
per component and to call the \ct{deployAll()} method to trigger the
deployment. The ROS middleware requires an implementation of the
\ct{NodeMain} interface. An extract of the deployment script for the
case study application is shown in Listing~\ref{listing:deploy}.

\lstinputlisting%
[float,language=java,%
caption={An extract of a developer-supplied Java deployment script for
  the case-study application},%
label={listing:deploy}]%
{code/Deploy.java}

\begin{figure}
  \centering
 \includegraphics[width=1\linewidth]{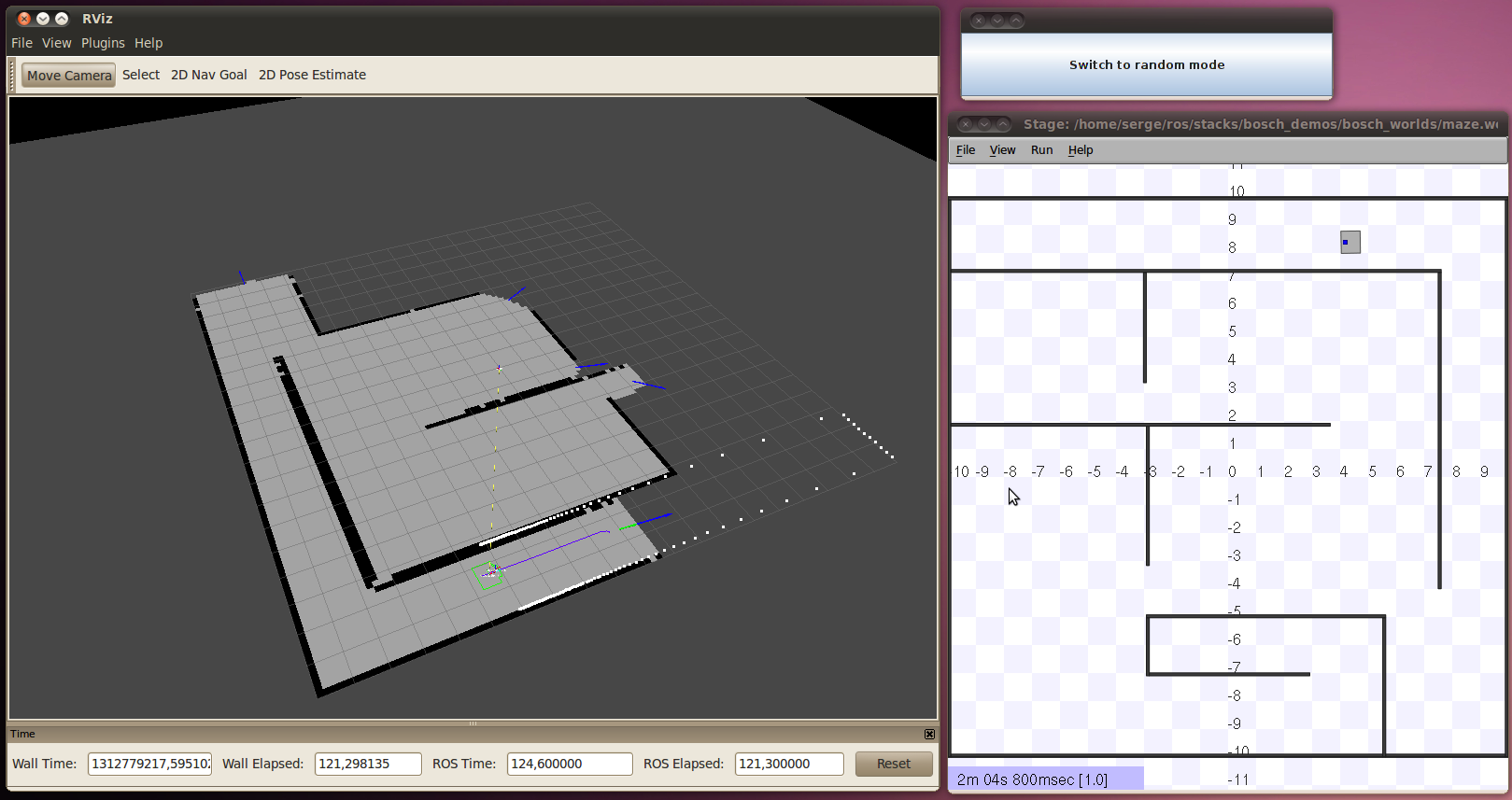}
 \caption{Screenshot of a simulation of the case study. On the left, a
   window displays the standard rviz visualization tool presenting the
   neighborhood visited by the robot in exploration mode. On the
   top-right, a button allows an operator to change the current mode
   of the robot. On the bottom-right, a window displays an instance of
   the Stage simulation engine}
\label{fig:diaspec-simulation}
\end{figure}

Figure~\ref{fig:diaspec-simulation} presents a running simulation of
our case study. The code generated is integrated in the ROS middleware
and the execution can be analyzed by the tools provided by ROS.

In this section we saw how to implement a robotics application on top
of a programing framework generated by the \diaspec{} compiler. This
programming framework calls developer's code when necessary and make
the development easy by passing everything the developer needs as a
parameter to abstract methods. In the next section we discuss the
benefits and problems of using \diaspec{} in a robotics setting.

%!TEX root=dslrob.tex

\section{Discussing}
\label{sec:discussing}

\diaspec{} decomposes the development of an application into two well
defined stages: a design stage for which \diaspec{} provides a
domain-specific design language and an implementation stage for which
\diaspec{} provides a design-specific programming framework. With the
design language and SCC architectural pattern, a developer is guided
in creating components with a single responsibility each, thus
enhancing reuse. An application design also explicits interactions
between components making the runtime behavior easier to understand.
With the programming framework dedicated to the design, a developer is
guided in creating an implementation for each component.
Indeed, the generated programming framework takes care of the control
loop of the application as well as all interactions between the
components. As a result, a developer can focus on implementing the
high-level application logic, letting the framework handle the
details. Moreover, the programming framework provides all necessary
pieces of required information directly as parameters to the abstract
methods. This reduces the amount of documentation required to start
using the programming framework.

In the previous sections we saw that \diaspec{} can be used to design,
implement and deploy a robotics application for a widely used
middleware. In the following we discuss various problems we have met
while applying \diaspec{} in a robotics setting.

\subsection{\diaspec{} Dynamicity}

\diaspec{} is capable of handling appearing and disappearing entities
at runtime. For example the following code lets the \ct{Motion}
context operator subscribe to sources of information from the
\ct{Exploration} and \ct{ModeSelector} entities:

\begin{lstlisting}[language=java,numbers=none]
@Override
protected void postInitialize() {
  discoverExplorationForSubscribe.all().subscribeTwist();
  discoverModeSelectorForSubscribe.all().subscribeMode();
}
\end{lstlisting}

This method has to be implemented in the \ct{Motion} Java class (for
the \ct{Motion} context operator). The programming framework takes
care of updating the subscription when a new entity appears or an
existing entity becomes inaccessible. As a result, a new exploration
mode or a new mode selector can be deployed at runtime. We believe
that in a robotics settings where most, if not all, entities are known
at deployment time this additional code is most of the time
unnecessary. Indeed, this code could potentially be inferred
automatically from the declaration of the \ct{Motion} context operator
and pushed inside the generated programming framework. However, in a
multi-robots settings, where a robot can discover services provided by
nearby robots, the \diaspec{} entity discovery and subscription
mechanisms could still be useful. The \diaspec{} design language could
be extended to let a developer declare which entities are known at
deployment time and which ones should be discovered at runtime. The
compiler could then leverage this additional information to generate
the necessary code in the programming framework thus reducing the work
required by the developers. We plan to investigate this issue in
future works.

\subsection{Data Type Reuse}

\diaspec{} allows the definition of new types (structures and
enumerations) as well as the importation of existing Java types. Very
often, a middleware such as ROS comes with its own data types. The
developer must then choose to reuse the data types coming from the
middleware or define new ones. Using the middleware data types can be
particularly useful as these data types can be complex such as the ROS
``twist'' data type. This is the solution we use for the case study
and the \ct{Twist} data type as is illustrated by the use of the
\ct{import} keyword in Listing~\ref{listing:design},
line~\ref{design:import}. However, choosing reuse of data types from a
middleware tightly couples the application with this middleware and
thus prevents potential for reuse of this application with other
middleware. Another solution is to develop new data types in
\diaspec{}. This makes the application independent from any underlying
middleware. However, this requires conversion code at the boundaries
of the application where communication with the middleware is
required. For example, it is possible to define the \ct{Twist}
data type within the design as follows:

\begin{lstlisting}[language=diaspec, numbers=none]
structure Vector3 { x as Float; y as Float; z as Float; }
structure Twist { linear as Vector3; angular as Vector3; }
\end{lstlisting}

Then, an implementation of the \ct{Wheel} entity would have to convert
from the \diaspec{} \ct{Twist} type to the ROS \ct{Twist} type:

\begin{lstlisting}[language=java, numbers=none]
private org.ros.message.geometry_msgs.Twist
                                   convert(Twist twist) {
  org.ros.message.geometry_msgs.Twist rosTwist;
  rosTwist = new org.ros.message.geometry_msgs.Twist();
  rosTwist.angular = convert(twist.getAngular());
  rosTwist.linear = convert(twist.getLinear());
  return rosTwist;
}
private org.ros.message.geometry_msgs.Vector3
                            convert(Vector3 vector) {...}
\end{lstlisting}

This solution makes the code harder to read and maintain. Moreover,
similar code has to be duplicated everywhere in the application where
a conversion is required. An intermediate solution is to develop new
data types in Java. This solution can embed required conversions in
the data type itself to avoid duplication. The resulting code is still
harder to read than the first one however.

\subsection{Decomposition grain}

During the development of the case study we noticed that following the
SCC architectural pattern and the steps proposed in
Section~\ref{sec:designing} resulted in fine grained components,
promoting reuse. It is however important that the developer pays
attention not to create too fine grained components which would make
the runtime behavior hard to understand and debug. Indeed, because the
generated programming framework handles the interactions between the
components, debugging very fine grained components requires stepping
often into the generated programming framework. This is cumbersome and
should not be needed. A possible addition to \diaspec{} could involve
a dedicated debugger which would let the developer debug his
application without stepping into the generated programming framework.

Even with a dedicated development environment, too fine grained
components make the system harder to understand. As a rule of thumb, a
developer can start by creating a coarse grained component and can
refine it when its implementation becomes complex, when the component
requires a lot of interactions with other components, or when parts
of its computation can be reused.

% LocalWords:  SCC hoc

%!TEX root=dslrob.tex

\section{Related Work}
\label{sec:related}

Several software engineering approaches have been proposed to lower
the complexity of robotics systems~\cite{Brug07a}.

\paragraph*{Middleware and Software Frameworks}

Numerous middleware and software frameworks have been proposed to
support the implementation of robotics applications (\eg{}
CLARATy~\cite{Claraty}, ROS~\cite{ROS} and
Player/Stage~\cite{Coll05a}). Such approaches attempt to cover as much
of the robotics domain as possible in a single programming framework.
This strategy often leads to large APIs, providing little guidance to
the developer and requiring boilerplate code to customize the
programming framework to the characteristics of the application. In
contrast, a \diaspec{}-generated programming framework specifically
targets one application, limiting the API to methods of interest to
the developers. Our code generator could potentially target these
middleware thus leveraging existing work and hiding their
intricacies from the developer.

\paragraph*{Component-Based and Model-Driven Software Engineering}

Component-Based Software Engineering for robotics
(\eg{}~\cite{Brug07b}) and Model-Driven Engineering for robotics
(\eg{} OMG RTC~\cite{OMGRTC}, SmartSoft~\cite{Schl09a}) relies on
general-purpose notations such as UML to model domain-specific
concerns. By using general-purpose and established notations, these
approaches leverage existing knowledge from developers and existing
tools. Even though such approaches propose a conceptual framework for
developing robotics applications, they only provide the user with
generic tools. For example, these approaches require developers to
directly manipulate UML diagrams, which become ``enormous, ambiguous
and unwieldy''~\cite{Picek08a}. In contrast, \diaspec{} abstracts away
such technologies, limiting the amount of expertise required from the
developers.

\paragraph*{Domain-Specific Languages}

\emph{Smach} is a Python embedded DSL based on hierarchical concurrent
state machines for building complex robot behavior from primitive
ones~\cite{Boren10a}. \emph{Smach} is tightly coupled with ROS, allows
only static compositions of behavior and can not adapt compositions to
new situations during execution. SmartTCL (Smart Task Coordination
Language) is an extension of Common Lisp that is used to do on line
dynamic reconfiguration of the software components involved in a
robot~\cite{Stec11a}: knowledge bases, simulation engines, symbolic
task planners, models and low-level hardware. At design time, the
developer defines execution variants that robot operates at runtime.
In order to lower robotics inherent complexity, analysis and
simulation tools could also be used at runtime to determine pending
execution steps with specific parametrisation before the robot
effectively execute them. Unlike these DSLs, \diaspec{} allows a
natural decomposition of applications according to the SCC
architectural pattern, guiding the work of the developer. Compared to
SmartTCL \diaspec{} lacks the ability to recompose the components at
runtime.

% Add to reference: David Kortenkamp, Reid Simmons, Chapter 8 - Robotic
% Systems Architectures and Programming, Handbook of Robotics, Springer
% Verlag.~\cite{Kort08a}

%!TEX root=dslrob.tex

\section{Conclusions and future works}
\label{sec:conclusion}

In this paper, we have proposed to use \diaspec{}, a domain-specific
design language for Sense/Compute/Control applications, in a robotics
setting. We have shown how this language allows a developer to
structure an application in fine-grained and reusable components by
following the SCC architectural pattern. Developing a complex
application shows the benefits of such an approach regarding reuse of
existing software components and diminution of complexity for the
developer. We have also highlighted  problems we have met
during the development of a standard robotics application.

Being able to adapt a robotics system to different capabilities and
resources is a key issue in software engineering for robotics. For
example, a robot can perform two similar missions differently with
different resources. Our approach facilitates changes to a robotics
system by making explicit the software components and their
interactions. However supporting static adaptation is insufficient in
a robotics setting as robots need to dynamically adapt to resource
evolutions (\eg{} failures and environment) while performing their
tasks. \emph{Resource-adaptive architectures} address dynamic
adaptations. However, such architectures are ad hoc solutions that
developers can hardly reuse and scale. Therefore, an ideal robot
control architecture should be \emph{resource-adapting}, \ie{} an
architecture that explicitly manages and represents
resources~\cite{Bour07a}. The main perspective of this work is to
introduce such dynamic variability inside \diaspec{}.

\section{ACKNOWLEDGMENTS}

The authors gratefully acknowledge the INRIA Phoenix research group
who granted the authors the authorization of use of DiaSuite, a tool
suite which includes \diaspec{}.

\bibliographystyle{plainurl}
\bibliography{dslrob}

\end{document}